Enhancing Law Enforcement Training: A Gamified Approach to Detecting Terrorism Financing

**Abstract**

Tools for fighting cyber-criminal activities using new technologies are promoted and deployed every day. However, too often, they are unnecessarily complex and hard to use, requiring deep domain and technical knowledge. These characteristics often limit the engagement of law enforcement and end-users in these technologies that, despite their potential, remain misunderstood. For this reason, in this study, we describe our experience in combining learning and training methods and the potential benefits of gamification to enhance technology transfer and increase adult learning. In fact, in this case, participants are experienced practitioners in professions/industries that are exposed to terrorism financing (such as Law Enforcement Officers, Financial Investigation Officers, private investigators, etc.) We define training activities on different levels for increasing the exchange of information about new trends and criminal *modus operandi* among and within law enforcement agencies, intensifying cross-border cooperation and supporting efforts to combat and prevent terrorism funding activities. On the other hand, a game (hackathon) is designed to address realistic challenges related to the dark net, crypto assets, new payment systems and dark web marketplaces that could be used for terrorist activities. The entire methodology was evaluated using quizzes, contest results, and engagement metrics. In particular, training events show about 60% of participants complete the 11-week training course, while the Hackathon results, gathered in two pilot studies (Madrid and The Hague), show increasing expertise among the participants (progression in the achieved points on average). At the same time, more than 70% of participants positively evaluate the use of the gamification approach, and more than 85% of them consider the implemented Use Cases suitable for their investigations. These outcomes are further discussed to detect the introduced approach's benefits and limitations and improve future events.

1. Introduction

Tackling terrorist financing through investigation, prosecution, and prevention is a worldwide issue that extends beyond Europe. Every day, terrorists find new channels to communicate, campaign and finance their activities. For example, as reported by EUROPOL in the IOCTA report (Europol, 2021) two main trends are related to crowdfunding campaigns and generating market revenue. In the first case, their schema is pretty easy: they raise a crowdfunding campaign to gather funds for their activities. In the second case, they try to generate revenue by selling extremist versions of common products or merchandise (such as Nazi-related items, ISIS promotional materials, etc.), as well as other legal and illegal goods (like counterfeit products, firearms, explosives, everyday items, etc.) to the general public or other extremists/terrorists. However, in both cases, to maintain anonymity, they often employ a combination of cryptocurrencies and markets in darknet technologies (Europol, 2022).

To tackle these needs and combat cybercrime, new paradigms, such as Artificial Intelligence (AI) and Big Data, are being used alongside conventional software to create novel investigation tools (Maher, 2017). For example, in (Mijwil, 2023), authors study emerging technologies such as ChatGPT in good practices and strategies to face cyberwar. However, these tools typically include multiple steps for collecting, processing, analysing and visualising information related to financial data (e.g., transactions, electronic invoices, etc.) and correlating them with context data extracted from social media analysis, forums, phishing acts, etc. (Kilger & Choo, 2022). This pipeline allows for improving the quality and the potential of the tools. However, they also enhance their complexity, requiring specialised domain expertise and technical profiles for their usage. As a result, Law Enforcement Officers (LEOs), Financial Intelligence Officers (FIOs), and even private financial investigators may be deterred from using these technologies (Klingberg, 2022).

In that sense, in several European (EU) projects, partners have tried to address these problems and increase engagement and technology transfer to Law Enforcement Agencies (LEAs) through different events like training sessions. For example, in the GRACE project (GRACE, 2020), LEAs are trained to use federated-learning AI tools to fight the propagation of Child Sexual Exploitation and Abuse (CSEA) materials. In the DANTE project (DANTE, 2018), similar structured training activities are organised with the aim of enhancing the analysis of terrorist-related content. In the i-LEAD project (i-LEAD, 2023), specific training sessions in using project technologies for tackling cybercrime and performing forensics investigations are planned, whereas, in CYCLOPES (CYCLOPES, 2023), events for training and testing participants on specific Digital Forensics tools are organised. Joint Live Exercises are explored in the CTC project (CTC, 2023) for providing both theoretical and practical knowledge about counter-terrorism financing. Field Labs events were organised in the TITANIUM project (TITANIUM, 2019) with the aim of training LEAs on studying criminal trade flow through cryptocurrency analysis. Yet, these events are based on CtF exercises, as well as happen for hackathon events deployed in the ASGARD project (ASGARD, 2016).

In this paper our research objective is to demonstrate that technical knowledge transference and student engagement can be achieved and evaluated in a context of high stress and dynamic prioritization typical in crime investigations. In particular, we propose the hypothesis that the research objective can be obtained by combining learning, training and gamification techniques in a unique methodology, and we present the deployment and results obtained in the Anti-FinTer (AFT) project (Anti-FinTer, 2023). The methodology combines traditional teaching techniques like lectures with moderated virtual learning environment (VLE), workshops and exercises with gamification techniques for facilitating interaction and engagement between participants. The aim of the AFT project is to train LEAs and Financial Intelligence Units (FIUs) to enhance their ability to use emergent technologies and complex pipelines to reveal financing activities of terrorism. In this way, it will be possible to increase Europe's ability to use novel tools for investigating terrorist financing and also promote EU technical and strategic sovereignty. For this reason, the AFT project exploits four tools that have been developed in previous EU projects, such as Graphsense (Haslhofer, et al., 2021) for virtual assets analytics, the Visual Analytics tool for forensic image processing, Ordainsare as a transaction anomaly detector based on the model presented in (Zola, et al., 2019) and the Dark Web Monitor[1] for analysing the darknet content.

The methodology followed in the AFT project is based on two main pillars: *Learning and Training* and *Gamification*. Learning and Training events are organised on four levels: *Asynchronous Courses, Knowledge Hub Meetings (KHM), Train-The-Trainers (TTT)* and *Face-To-Face (FTF) training* with different aims and target groups. On the other hand, the gamification task is based on hackathon events. These events are designed as Capture-the-Flag (CtF) exercises that allow participants to learn effectively how to use AFT tools (alone and linked) in their day-to-day work for revealing terrorist financing activities.

All these activities and events have been designed as experiments, following an iterative approach that allows temporal comparison and improvement validation and defining the generation of evidence during the whole learning and training process (completion measures for the objective indicators and satisfaction for the subjective ones).

---

[1] https://cflw.com/dwm/

The restricted access for practitioners, determined by the number of events and participants in each, has led to the application of the entire methodology to all. The availability of practitioners to participate has been the primary constraint we need to manage, resulting in participation disparities in each learning activity. These restrictions prevent a deeper ablation study that would allow us to compare the impact of each learning/training activity over the overall goal.

Although in this study we present the whole methodology, the main analysis is performed on the outcomes achieved during the Asynchronous Courses, as well as the general results obtained in two AFT FTF training and hackathon events, the first held in Madrid in 2022, and the second held in The Hague in 2023. In each of these two events, AFT stakeholders and external users were involved in two days of activities. The results indicate a satisfactory level of engagement among the participants for both sessions, and attendees positively valued the general AFT contents and tools. However, they showed doubts about their use in day-to-day investigations. This analysis also helped identify limitations and organisational weaknesses that will be addressed in preparation for the future (and last) AFT FTF training and hackathon event scheduled in Vienna at the end of the project. The obtained results and findings can be easily transferred to learning and training related to other cybercrime investigation domains that share the same tools and similar methods, like Child Sexual Exploitation and Abuse, Weapon/Drug Trafficking, Money Laundering, Ransomware or Political Corruption.

2. **Background**
    a. **Andragogy and Pedagogy training**

Training participants are, in general, experienced practitioners in professions/industries that are exposed to terrorism financing, and all arrive with a preconceived concept of the topic. In line with the expected learning outcomes of the project, we must incorporate a central objective of understanding foundational and prevailing knowledge associated with terrorism financing and adjacent issues while providing participants with the autonomy to engage with learning materials they feel best suit their needs. This approach is critical with the target of educating adults (andragogy). The andragogical response to the training structure must leverage the learning opportunities provided by these activities while also appreciating the project participants' diverse skill sets. In this sense, the main objectives of the training structure can be summarised as follows:

i.  Appreciate the requirements of the adult participants;
ii. Provide alternative routes of learning within the materials;
iii. Generate a learning environment rooted in experiential learning;
iv. Provide knowledge that can be perceived as immediately applicable or useful;
v.  Create a learning environment that can leverage the insights of training participants.

   i. **Applicable Theories of Learning**

Given the limited time available to disseminate the learning materials, an instructor-led approach on its own may not be conducive to achieving satisfactory training outcomes. Terrorism Financing and its associated issues is a complex and ever-evolving topic requiring intricate insight and knowledge across various subjects. An instructor-led approach for this environment would necessitate the use of behaviouristic learning materials, where instructors condition learners to reach a preconceived standard of competency through reinforcement, repetition, and variation (Carlile & Jordan, 2005).

Achieving behaviouristic learning outcomes while adhering to the training structure objectives may be impractical.

It is important to leverage the prior knowledge and the subject area of interest expressed by the participants and use this as the central focus of the teaching environment. This approach allows the production of deeper insights over the short engagement period. In this sense, it is important to diminish the instructor's role as the exclusive provider of knowledge and reallocate a degree of control and accountability to the learners. A number of learning theories are compatible with this type of learning environment while still adhering to the objectives. Nevertheless, certain elements of the humanist approach need to be incorporated into the training structure. Autonomy is provided to participants to learn independently and without the burden of stress typically associated with assessment structures. The intention behind this approach is to instil a deep sense of personal appreciation for the subject matter, such that the participants will desire further learning following the cessation of the training (Governors Western University, 2020).

There is a need to account for the limited contact hours that instructors will have with the participants. A lack of consistent contact hours can lead to a disjointed base of learning for participants, limited engagement with the learning materials, and suboptimal learning outcomes. Therefore, creating an empathetic and shared social environment is paramount to achieving the expected learning outcomes. The *social constructivist* learning environment leverages the diverse array of knowledge and experiences offered by each learner to develop social discussions within the learning environment. In addition to constructing their own view on the knowledge provided by the instruction materials, participants can reach a higher plane of understanding by simultaneously sharing their experiences, perspectives and concerns with peers on the subject matter being addressed (Topping, 1998). Developing an interaction-based environment has been heavily linked with student learning and satisfaction. Furthermore, instructor and peer dialogue has been shown to develop trust and social interactions in learning settings (McLean, 2018).

### ii. Achieving Practical Learning

The teaching philosophies identified as suitable for the AFT training structure (humanism, social constructivism) can inform the appropriate learning structure to implement within the learning environment. A key aspect of this is identifying that training will take place both online and in a face-to-face environment. The andragogical aim of the face-to-face learning environment is to establish a community of inquiry, an andragogic framework where learning occurs at the intersection of social, cognitive and teaching presence (Garrison & Arbaugh, 2007), and a community of practice where knowledge is embedded in the activities, social relations and expertise of specific communities (O'Neill & McMahon, 2005). Pre-prepared learning materials (cognitive) are paired with synchronous discussions on the topics covered (social). The structure is designed to foster individual perspectives on issues related to Terrorism Financing (cognitive) and encourage interactivity and feedback amongst peers as a means of constructing new insights and perspectives (social). These will be underpinned by specific prompts that will be introduced to the learning environment to spark discussions among the participants. Moreover, the identification of a sufficiently motivating problem that serves as a platform for investigation is important in inquiry-based learning environments (Finkel, 2000). In addition, the practice-based learning environment is incorporated into the learning environment since all participants are stakeholders in investigating terrorism financing. While inquiries will provide valuable insights, introducing case study examples of terrorism financing issues can lead to practical, ready-to-apply knowledge. Introducing case studies to explore is particularly effective in enhancing learning outcomes when real-world problems remain unresolved and ill-structured (Barrows, 2002). Once this problem is introduced, the subsequent investigation and discussion is where the learning takes place.

Participants will be highly intrinsically motivated to learn what is necessary to solve it (Auman, 2011), given that counteracting terrorism financing and discussing the issues of the prevention of terrorism financing, is a common interest shared by all participants.

### b. Gamification

Capture-the-Flag (CtF) is a type of information security contest where participants are challenged to solve a range of tasks in order to obtain a designated item called a flag. At a high level, there are many flavours of cybersecurity competitions, as well as platforms for managing them. In this sense, it is also difficult to define a finite set of CtF strategies. Several studies (ENISA, 2021), (Švábenskỳ, et al., 2021) define two main CtF strategies: *Attack/Defend* and *Jeopardy-style*. In particular, the lack of attack/defend team specialization, the presence of multiple users, and especially the structured learning goals of the AFT project led us to choose the Jeopardy-style format to create a competitive environment and engage multiple users simultaneously.

The CtF strategy has resulted in wide success in terms of introducing and learning cybersecurity-related concepts (Švábenskỳ, et al., 2021), but also motivating continued learning after the exercise (McDaniel, et al., 2016). For example, in (Huang, et al., 2011), CtF is used for solving as a differential game, whereas in (Eagle & Clark, 2004), this strategy is used to educate students to act as crackers and find new vulnerabilities in existing systems (data, files, devices, etc.). Another interesting work is presented in (Chicone, 2020), where authors compare two CtF platforms (Facebook CtF and CtFd) for cybersecurity learning. In (Hanafi, 2021), the authors study how gamification can be applied to introduce students in Cybersecurity Education, who are, in fact, non-technical backgrounded targets. As already mentioned, this gamification approach is also used in many EU projects for improving knowledge and technology transfer.

Inspired by these previous works, in this paper, we combine learning and gamification strategies for facilitating knowledge sharing and best practices exchange among AFT stakeholders and end-users. This approach has shown to be helpful in improving law enforcement capacity and developing expertise in using emerging AI tools for terrorist financing investigations.

### 3. Methodology

Designing curricula and training programs for crypto-asset analysis technologies is challenging due to their intricate and rapidly evolving nature. In this sense, in the ATF project, we combine both asynchronous and synchronous learning in order to maximise the impact of the activities and reach a broader audience (Figure 1). More specifically, we started designing an *Asynchronous Course* deployed in a Learning Management System (LMS), with the aim to uniform attendees' knowledge related to the general topics treated in the project, such as crypto finance, financial regulation, dark web structure, crypto ecosystem, machine learning for investigation, etc. On the other hand, three different activities are taken into account for synchronous learning, such as Knowledge Hub Meeting (KHM), Train-the-Trainers (TTT) and Face-To-Face training (FTF training), as shown in Figure 1. Finally, the AFT training methodology also includes gamification (*Hackathon*) of domain tasks for practising the acquired knowledge. Although in the following sections, we describe all the activities, this study mainly focuses on presenting the results obtained from three of them: Asynchronous Course, TF training and Hackathon.

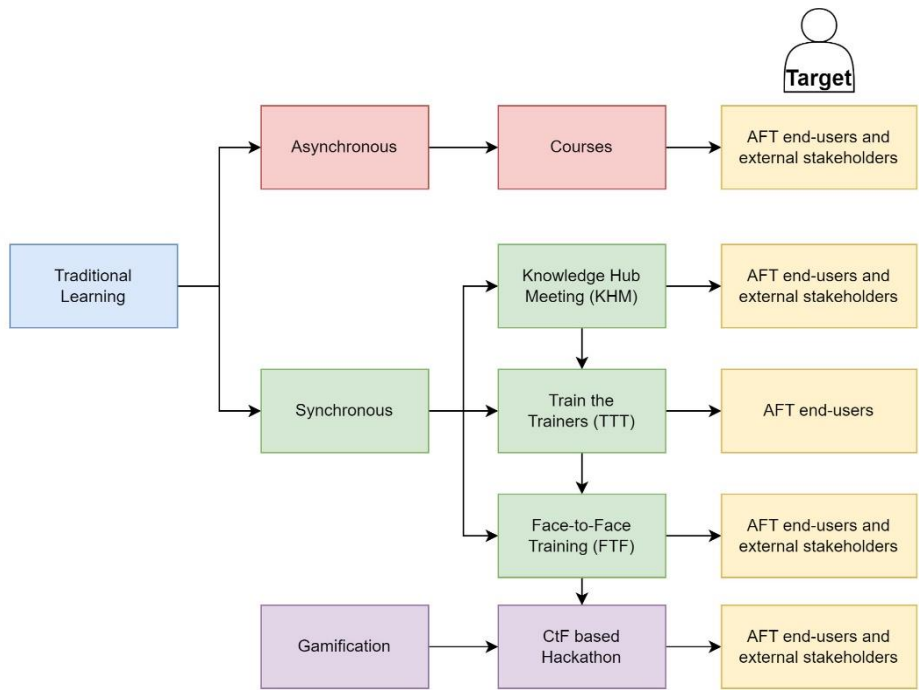

*Figure 1*

a. **Asynchronous Course**

Given the profile of the AFT participants, embedding flexibility was at the forefront of the programme design and considerations were taken in leveraging key dimensions of learning and teaching. There are several principles and established pedagogical approaches applied, appropriate both to the discipline and to the level of the award. An asynchronous, active online learning approach underlies the conception of teaching and provides a strategic filter through which the course is shaped and supported. AFT end users are post-experienced, in full-time employment and are viewed as not just participants but as active learners and critical thinkers with their own emerging theories about the world. This view also supports the principal tenets of adult learning in that they desire and enact a tendency toward self-directedness, practical learning (tools usage), experiential learning, and a flexible learning environment. This relative informality is not without structure, and clear expectations are set regarding performance and rules of engagement as this learning journey is one embarked on together.

In fact, participants' own experiences are an essential resource for learning. In particular, given their rich and diverse backgrounds, they bring unique perspectives and insights to the course, creating fertile ground. Thus, considering their experiences as valuable input, the AFT course design and delivery promotes a process of interaction between what is known and what is to be learned, as a reflection on experiences and understanding is the foundation upon which something new is built. By seeking participants' points of view and using this to establish their current conceptions, the objective is knowledge construction (and sometimes reconstruction) rather than knowledge reproduction.

The asynchronous AFT course, hosted on the Learning Management System (LMS), comprises 12 lectures organized into two parts over a six-week period. Part One (Lectures 1 to 6) focuses on understanding cryptocurrencies, their operations, and their impact on regulatory, law enforcement, and tax authorities. Conversely, Part Two (Lectures 7 to 12) delves into technical concepts utilised by AFT tools like Machine Learning, Image analysis, Follow-the-Money approach, and darknet, alongside tool presentations. Lecture 12 is a reflective session, gathering feedback and discussing tailored learning paths for different organizational needs.

Pre-requisites for effective teaching in this regard are interaction, dialogue, and reflection. For this reason, a considered level of meaningful interactivity is designed into the AFT course to encourage participants to critically engage and enhance the process of cognitive development that results in more significant evidence of the achievement of learning outcomes, as this allows for deeper engagement with concepts as multiple perspectives challenge existing assumptions. This interactivity has been provided using the Discussion Forums added to each subject, and on the other hand, with programmed live sessions to ask directly to the subject coach. The research method used to evaluate knowledge acquisition is the results of short quizzes related to each subject of the course and the final completion of all of them as the main indicator.

### b. Knowledge Hub Meeting (KHM)

Knowledge Hub Meeting (KHM) is a pan-European and multi-disciplinary knowledge hub in which end users: LEOs, tax authorities, financial services institutions, but also academia, the private sector and any interested stakeholder can participate to discuss best practices and policy recommendations about new crypto-threats, investigation trends, and novel detected modus operandi. This knowledge-sharing approach is implemented through a series of at least 20 workshops (physical and online) in two years, which increases the attendees' expertise and helps the technical partners understand the needs of users and stakeholders, allowing them to improve their tools. Therefore, the KHM aims to create a community (long-term goal) supported by the European Anti-Cybercrime Technology Development Association (EACTDA) to keep up with the developments beyond the lifetime of the project. This activity has been designed as an open discussion and network creation among practitioners, and the engagement has been demonstrated with the attendance at each event.

### c. Train-the-Trainers (TTT)

Train-The-Trainers (TTT) event is an online training that involves developing the skills and knowledge of individuals who are responsible for training others. This activity has been initially focused on technical trainers, but it is now open for end user trainers. More specifically, TTT is carried out with two main goals; on the one hand, it presents how the AFT tools can enrich the traditional investigation methods in a complementary way. On the other hand, it shows the participants how to use appropriate teaching and good practice for coaching, in turn, colleagues and other organisation members. In fact, participants' existing knowledge and their interest in the subject area become crucial, serving as the pivotal point around which the teaching environment revolves. This approach aims to enhance the quality of the insights within the limited duration of engagement. In that sense, we removed the instructor as the sole purveyor of knowledge and transferred an extent of control and responsibility into the hands of the learners. The instructor's primary focus in this approach is to facilitate learning and promote participant engagement rather than act as the sole source of knowledge. This shift from instructor-centred to learner-centred teaching requires an environment that encourages autonomy in constructing knowledge, active participation and knowledge sharing among the participants. The research method used to evaluate the improvement in knowledge transference is the results of the survey related to trainer evaluation completed after each FTF training and Hackathon event.

### d. Face-To-Face (FTF) training

Face-To-Face (FTF) training events are structured to provide a comprehensive explanation and practical demonstration of the AFT tool functionalities while ensuring participants are not overwhelmed by an excessive influx of new concepts. In this way, the AFT end users and external stakeholders who take part in the hackathon event are trained for the next day when they effectively use these tools (alone) to address realistic challenges.

Both Madrid and The Hague FTF events adhered to an identical schedule. It comprised six 30-minute sessions, totalling 3 hours. The sessions covered introductory domain concepts on Crypto Governance and Regulations, followed by presentations of the four AFT project tools. Each tool was showcased through a 20-minute demonstration and a 10-minute Q&A session. The final segment introduced an integrated platform, complying with AFT project specifications, enabling swift access/switching between tools.

The choice of these temporal slots allows us to ensure enough time for presenting each tool and have a demonstration of it but also avoid overloading the participants with too many new concepts. The research method used to evaluate knowledge acquisition has been included in the survey provided after each FTF training and Hackathon event.

### e. Hackathon

AFT hackathon events are based on the gamification of investigations related to terrorism financing activities, and it is oriented toward FTF-trainer users. This is performed using a Capture-the-Flag (CtF) structure and deploying different domain challenges following the Jeopardy-style format, where participants should provide a response in the shape of a "tag" for each question presented and where the difficulty is increasing step by step. In particular, to accomplish the AFT project goal, the challenges are aligned with possible cybercrimes such as financing terrorism, money laundering and fraud detection that involves Dark Web, crypto-assets, new payment systems and darknet marketplaces. In this sense, AFT technical partners are responsible for designing realistic challenges based on the *modus operandi* analysed during the project. However, they also need to consider that hackathon participants are not familiar with these new AFT tools, so challenges need to be created accordingly. For this reason, all along the AFT project, different strategies are drawn and deployed to create the challenges and to guide the users through the tools, diminishing the constraints and tool-related hints for specific tasks gradually. More specifically, we started the project with a more structured and guided approach (Madrid hackathon), then proceeded to lessen constraints and limitations about the tools (The Hague hackathon) to finally enable participants to attain full independence in choosing the appropriate tools for the right tasks (final event).

Regarding the structure of the hackathon, each event is separated into sessions, in each of which participants are required to tackle progressively more challenging tasks, beginning with the exploration of basic functionalities and operations and culminating in demonstrating the true capabilities of AFT tools and their practical relevance in real investigations and complex scenarios. Thereby, this approach promotes involving AFT users and external stakeholders, creating a more competitive playground, and allowing AFT technical partners to speed up development/validation cycles with respect to traditional innovation processes.

To improve the learning process, each hackathon event follows a different strategy in the challenge definition. In particular, the first hackathon adopted a tool-centric strategy, whereas the second implemented a challenge-oriented strategy. Consequently, these strategies influence the duration of each session and the overall duration of the whole event.

On the one hand, the tool-centric approach (Madrid hackathon) included separate sessions for testing each AFT tool with easy challenges to aid user familiarity, guide actions, and assess usability. Tasks were self-explanatory, ensuring clarity and minimizing errors. Over five hours, distributed in 60-minute blocks, participants progressed from basics to more complex challenges requiring deep expertise.

On the other hand, the challenge-oriented strategy (The Hague hackathon) began with a tool familiarization session for both new and returning participants, as planned in the Madrid event. However, this time, instead of splitting challenges per tool, they were consolidated to accelerate learning. Subsequent sessions targeted a real-world scenario: addressing terrorism financing via the Luckp47 darknet market, which is a niche darknet market that allows the purchase of various illegal items using Bitcoin and is purportedly affiliated with a paramilitary group (Jiang, et al., 2021). The challenge-oriented ended with a session in which participants were asked to use the acquired knowledge to address more complex scenarios, which could also involve the usage of more than one tool at once (similar to the last session in the Madrid event). In The Hague event, each session required considerably more time to complete, and as a consequence of this approach, the event extended beyond 5 hours.

Finally, both events ended with a timeslot for awarding the participants with the highest score and for collecting feedback.

The research method used to evaluate the knowledge acquisition has been gathered in the FBCTF tool using the challenge-solving metrics to analyse the completion, the fails, and the speed of challenge-solving. The number of completed challenges is also a measure of engagement with the content and tools provided. The survey provided after each FTF training and Hackathon event allows us to evaluate the supporting activities to improve them in the next hackathon.

### f. Study Methodology

Each learning and training activity has some type of evaluation metric to gather, enabling the generation of more robust engagement and validation of knowledge acquisition. These metrics are used as indicators and reflected formally in AFT deliverables. Those results are shared between the technical partners of the project, allowing them to improve the learning and training material available for the next synchronous event (Madrid, The Hague or even the final one in Vienna). Although the number of participants is variable in the synchronous learning events, and their workload changed along the time in the asynchronous ones, normalization of gathered results and trend analysis allows us to validate improvement in knowledge acquisition and engagement.

Furthermore, it is to be noted that for the guiding principle for both the design of tools for law enforcement and the protection of research subjects, the project was aligned with the ALLEA framework[2]. This European research integrity framework allows the creation of a strategic response to professional, legal and ethical responsibilities and acknowledges the importance of the institutional settings in which research is organised. For example, for gathering feedback at each event, participants were informed about the anonymity of the survey and the voluntary involvement.

## 4. Asynchronous Course Validation
### a. Validation context

All lectures are fully online, asynchronous and moderated by the course leaders. Discussion forums, with prompting questions, are embedded in the design to encourage collaboration with peers and course leaders and create a community of practice, drawing on the different backgrounds and experiences of the participants. Each week, course leaders hosted an online synchronous session to address specific questions and to further engage with participants. In this sense, participant engagement with assessment activities is a requirement of the course to ensure that learning outcomes are met. A reflective space is provided, allowing participants to evaluate their own learning experience and critically reflect on the content each week. Learning can be complex, and each follows its own strategy to navigate and engage with the content. For this reason, it is important to provide various motivational aspects and a multi-modal approach to deliver support to each participant in their learning journey.

---
[2] https://allea.org/horizon-europe/

Each lesson, excluding the last one, requires the completion of components (e.g., quizzes) to demonstrate participant achievements and engagement level but also for tracking features and evaluating the learning outcomes. At the same time, on achievement of proficiency levels, the next course became available. Completion reporting is also used to track activity for each course requirement in the programme and award a certificate to successful participants.

### a. Results and Lessons Learned

Figure 2**Error! Reference source not found.** reports the results in terms of the academic performance of the group and their progress through the various elements of the VLE. The figure shows a strong and consistent engagement throughout the course, although there was some attrition over the six-week duration. This was gradual, and no one element of the course caused a precipitative fall in the numbers staying on the course. Of the 81 who joined us at the start of the course, 52 personnel from across the European Union completed all 11 units. Furthermore, excluding Lecture 1, in all the others, the number of participants who passed the quizzes is very similar to the number of course participants, i.e., almost all the participants were able to complete the lessons successfully.

A completion rate of more than 64% (i.e., 52 out 81) for an online offering aimed at busy professionals is to be viewed as a success. Feedback was positive, and engagement on the discussion boards was reasonably strong. In fact, the discussion boards allowed the creation of a community of practice amongst participants and helped to have a direct chat with the teachers/moderators. Furthermore, positive feedback was also obtained about the structure of the course, its context, and technical content.

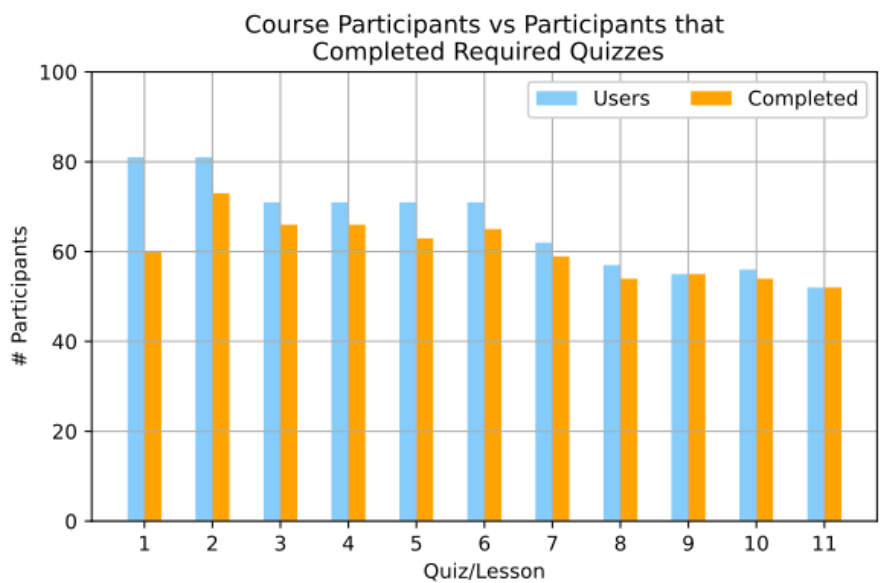

*Figure 2*

## 5. Face-To-Face Training Validation
### a. Validation context

FTF events are targeted towards the hackathon participants, which are mainly AFT end-users and external stakeholders from LEAs and FIUs. For this reason, to evaluate the events, a unique survey with 10 questions is prepared. The first 5 questions are related to the satisfaction grade of the training materials, lessons learned, and understandability of guides and platforms. These questions are designed to receive responses on a five-point Likert scale ranging from *Strongly Disagree* (1) to *Strongly Agree* (5). Then, the last 5 are open questions related to the most engaged and distanced training moments and which training actions they consider most helpful or puzzling.

For the sake of simplicity and the aim of this study, just the first 5 questions are reported and discussed. Indeed, the results from the open-ended questions hold significance only within the context of the project rather than being pertinent to the presentation of the methodology's advantages and limitations.

### b. Results and Lessons Learned

In the Madrid event, 15 of the 17 participants filled out the survey, whereas in The Hague, 13 out of 15. This number of participants can be considered limited to provide an extensive validation of the methodology. However, our idea is to follow and analyse the trends of the participants during the different events in order to check their learning progress.

Participants showed very broad opinions about how the knowledge gained in the training event is helpful for their immediate (short-term) investigations (Figure 3A). In fact, about 50% of participants in both events *strongly* or *simply* agree on that, whereas about 26% (Madrid) and about 15% (The Hague) have a different vision and disagree. The rest of the participants do not have a clear opinion. Different results are instead obtained by looking at Figure 3B, i.e., analysing how the training information would be beneficial to their day-to-day work in a long-term period. In this case, in the Madrid event, just 66.7% of the participants *strongly* or *simply* agreed with this claim, whereas in The Hague, all the participants expressed certainty in that regard. A similar trend is also followed in relation to how participants perceive the engagement and the obtained benefit of other players (Figure 3C). Finally, another interesting feedback is obtained with respect to the quality of the learning materials (guides, platforms, leaflets, etc.). In fact, Figure 3D shows that in the Madrid event, although a significant portion of participants (66.7%) endorsed the thoroughness and comprehensibility of the materials, 20% expressed dissent. However, in the second event, only 7.7% of participants assessed the material negatively, whereas more than 80% considered it valuable. The results presented clearly indicate that, on the whole, the AFT FTF training events receive a positive assessment from end-users and stakeholders. These users also express a sense of engagement in acquiring domain-specific and technical knowledge relevant to their daily responsibilities. Their recommendations were fundamental to detecting limitations and situations to be refined. In this sense, the iterative approach followed in this project contributes to a higher level of satisfaction, as evidenced by the enhanced outcomes witnessed in the second event.

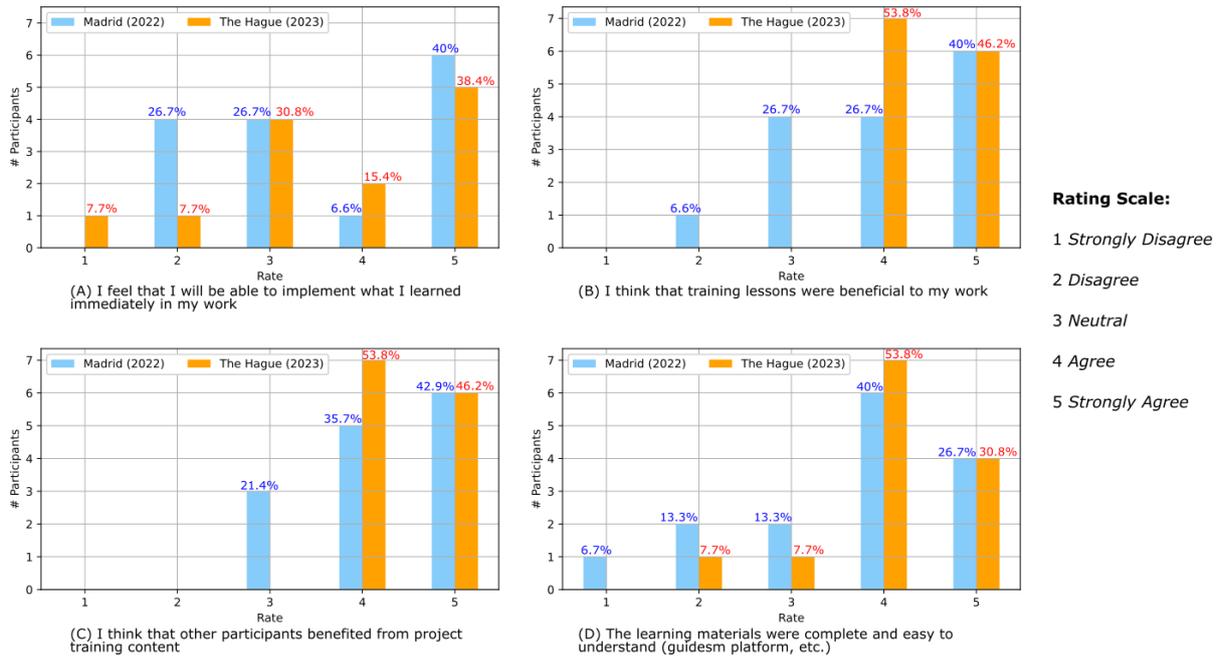

*Figure 3*

### 6. Hackathon Validation
#### a. Validation Context

The feedback is gathered just after the awards ceremony, using surveys, and it has its own timeslot, searching for a high engagement. First of all, they are collectively informed about the anonymity of the survey and the voluntary involvement. Then, the survey is delivered covering questions about the events, the organisation, and the tools. More specifically, opinions from AFT end users and external stakeholders as well as from AFT technical partners were gathered.

**AFT technical partners:** A specific questionnaire is prepared for them, consisting of two main sections: *exercise preparation* and *performance evaluation*. The former includes 10 questions related to the materials, location, facilities, complaints and problems in the hackathon organisation. The latter contains 15 questions separated into four different topics: tool installation (8 questions), configuration/default parameters (2), integration with other project tools (3) and data evolution (2).

**AFT users and external stakeholders:** Two evaluation frameworks are used for collecting this feedback, one based on the common state-of-the-art *(a) learning schemes* and the other based on *(b) objective metrics*. In the (a) case, a survey with 35 questions is designed. This questionnaire is created following two distinct learning schemes: the System Usability Scale (SUS) (Peres, et al., 2013) and the Kirkpatric Model (Smidt, et al., 2009). The SUS is composed of 10 standardised questions used to assess the usability of a wide range of systems. These questions are rated on a five-point scale ranging from *Strongly Disagree* (1) to *Strongly Agree* (5). On the other hand, the survey is improved with 25 (five-point Likert scale and open) questions based on the Kirkpatric model. This model comprises four criteria levels: Reaction, Learning, Behaviour, and Results (Bates, 2004). The first level aims to gather information on whether the participants perceive the task as relevant to their daily job or not. The second level assesses the knowledge acquired by the participants during the hackathon. The third level focuses on evaluating the change in the participants' attitudes and behaviours resulting from the newly acquired knowledge, which is a critical step as it can be challenging to validate effectively. Finally, the fourth level directly evaluates the performance results obtained during the development of the hackathons. In the (b) case, regarding objective metrics, data and logs are directly extracted from the FBCtF platform and analysed to evaluate the trends and statistics of each participant as well as their engagement level. This analysis gives us an overview of the difficulties and problems encountered by each participant, and, at the same time, it helps us to determine the winners of the hackathon event. Yet, the results allow us to fill the gap between the perception of trainees about their learning process and their actual performance.

It is important to note that although the objective metrics - (b) case - carry relevance within the scope of the project, they do not directly influence the presentation of the methodology's strengths and weaknesses. Therefore, in this work, we just report the number of participants who have correctly addressed each challenge and their final score to show their engagement level in the events. On the other hand, considering the extensive number of questions (35) in the survey - (a) case - only the ones related to the evaluation of the general methodology are further analysed and discussed.

b. **Results and Lessons Learned**

During the Madrid event, the survey was completed by 14 out of 16 participants, while in The Hague event, it was filled out by 11 out of 15 attendees. As previously pointed out, due to the limited number of participants, the main idea of this work is to monitor the participants' progress across various events to track their learning trajectory and the benefits and limitations of the proposed methodology.

Figure 4 reports the distribution of the expertise of the participants in both events. The figures show that, in both events, the number of officers that work directly on real investigations (operational level) overwhelms the number of strategic ones. In fact, in both cases, operational officers represented more than 75% of the attendees, although, in the Madrid event, there were more FIU agents with respect to LEAs. These outcomes align with expectations, as the AFT objective is to showcase the effectiveness of innovative tools for detecting terrorism financing primarily intended for operational officers' usage.

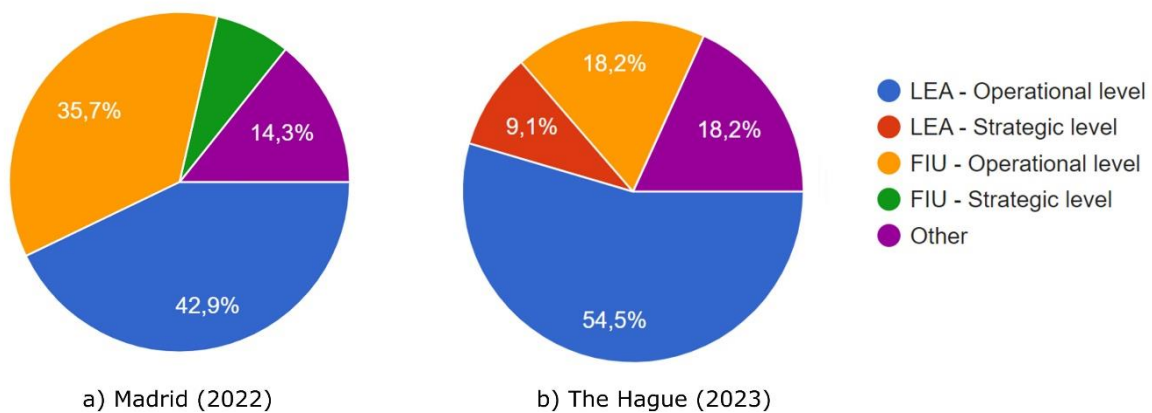

a) Madrid (2022)   b) The Hague (2023)

*Figure 4*

According to the information gathered in The Hague event (second hackathon), a total of 7 participants out of the 11 that completed the survey (equivalent to 63.7%) were also present at the Madrid event (first hackathon). As a result, three comparative questions were exclusively directed toward this subgroup. In particular, all of them positively evaluate the assessment and the improvements made in this second event, especially the new format of the hackathon based on a *challenge-oriented* strategy, the new functionalities introduced in the tools, and the design of the challenges related to a real investigation (Figure 5).

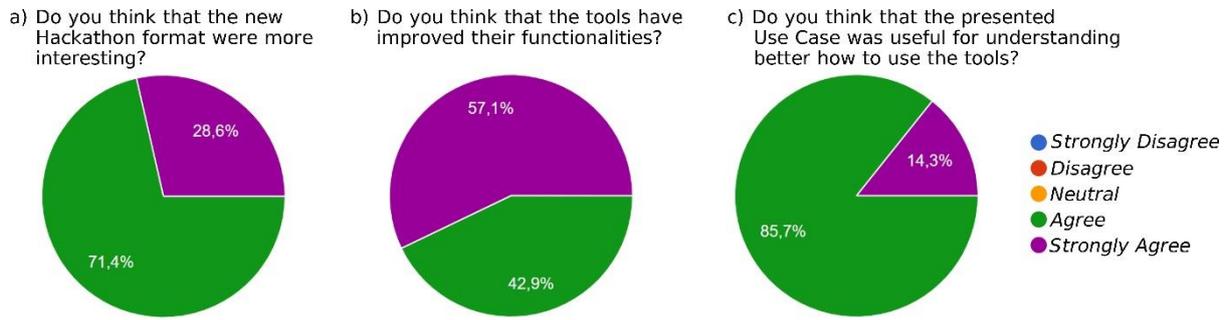

*Figure 5*

Figure 6A and Figure 6B show that all the participants felt confident with the technical and domain information acquired in the second event. In the Madrid hackathon, two participants considered the acquired domain knowledge insufficient, while one participant expressed the same opinion regarding the technical aspects. At the same time, four (domain) and five (technical) participants did not express a clear opinion (neutral). On the other hand, in the second event, just one participant was neutral about the tool knowledge, and all the others appreciated the gained information in both aspects. Regarding the usefulness of the concepts learned during the hackathon, participants still found issues understanding how they can be integrated into their daily duties (concrete use cases). In that sense, Figure 6C shows that about 47% and 64% of the participants - in the Madrid and The Hague events, respectively - did not consider these lessons learned relevant or they were neutral. On the other hand, if they were asked about their perception of using the knowledge in real investigations, i.e., with a more general point of view and not only related to specific day-to-day use cases, they show a different opinion (Figure 6D). In fact, especially in the second event, more than 80% of them had the perception that these lessons learned could be applied effectively in practical investigations.

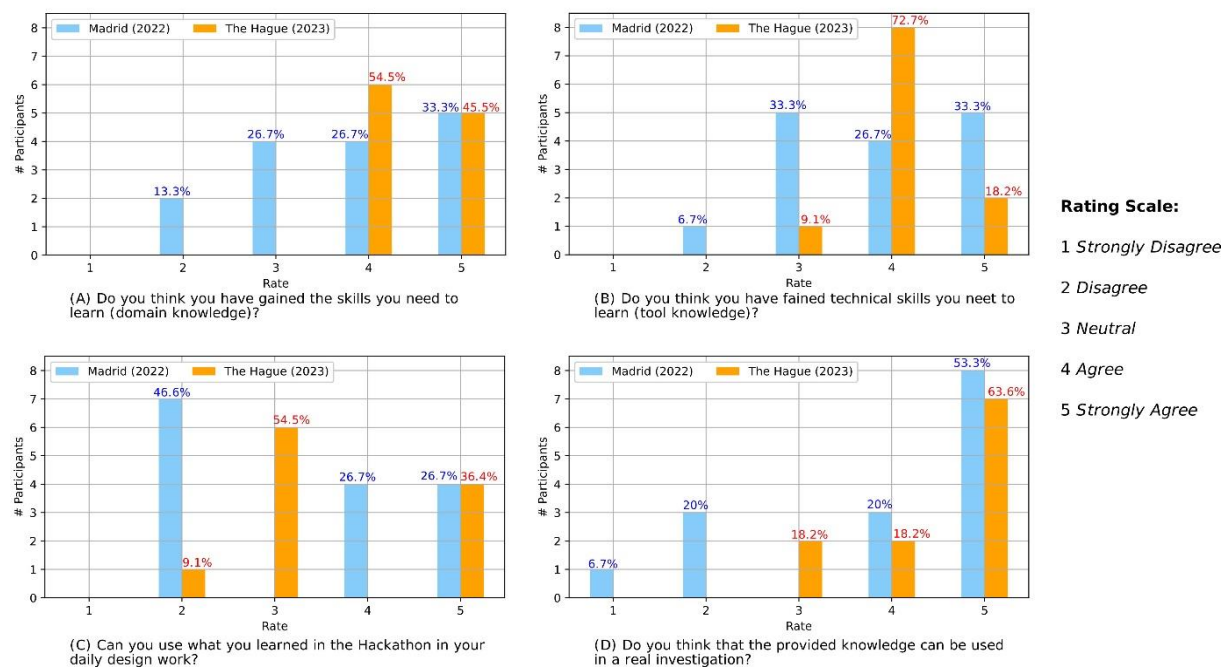

*Figure 6*

In terms of the engagement level of the participants, evaluated using objective metrics, the Madrid event achieved relatively homogeneous results. Figure 7 (left) shows that four players (25%) achieved extraordinary scores of more than 2,000 points out of the 2,850 available (red dotted line), and other eight players (twelve in total, 75% of participants) achieved more than 1,425 points (black dotted line). Yet, two more players reached at least 1,000 points, whereas only two were below this last threshold. For almost all the participants, Session 2 was the one in which they earned more points, as expected since it had a longer duration and involved two AFT tools. On the other hand, in The Hague event, a different trend is observed (Figure 7 right). In fact, despite the good feedback and comments gathered with the survey, only one player completed all the tasks and nearly attained the highest possible score, while six other additional players (less than 50% of the attendees), were able to reach at least half of the maximum score (black dotted line).

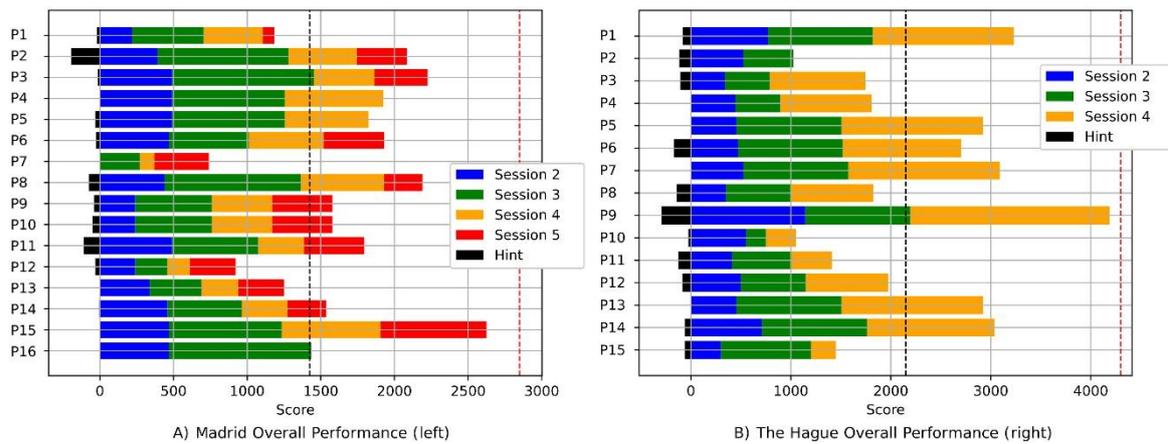

Figure 7

7. **Conclusion**

This paper describes a framework for training Law Enforcement Agencies and Financial Investigation Units to enhance their ability to use emergent technologies and complex pipelines to reveal financing activities of terrorism. This methodology combines learning and training with gamification activities. The first ones are organised to facilitate the exchange of information related to new criminal payment systems and modus operandi among different stakeholders, whereas the latter are designed as Capture-the-Flag exercises, allowing participants to learn effectively how to use AFT tools in their day-to-day work. The goal is to enhance the number of engaged participants coming from different fields and with different expertise.

The methodology was evaluated in two events held in Madrid (2022) and The Hague (2023). The asynchronous course, FTF training, and hackathons validate knowledge transfer through quizzes, contest results, and engagement metrics like quiz completion (over 60%, gradually declining in the 11-week course). There's an observed progression between events, averaging 1500 points in Madrid and 2000 points in The Hague. The overview of the results presented shows a satisfactory level of engagement among the participants in both training and hackathon, as can be seen in Figure 5, where 71.4% of the participants agreed that the Hackathon format was interesting, and 85,7% thought that the raised UCs were useful. Yet, attendees expressed a favourable view of the general AFT content and tools and recognised the potential of the acquired knowledge (domain and technical) for their investigation. Metrics in Figures 6A and 6B are proof of that, where data in the histogram shows that participants felt confident. On the other hand, they highlighted hesitation in using them in their day-to-day duties, as shown in Figure 6C. For this reason, as a lesson learned, AFT technical partners need to spend more time understanding the day-to-day needs of the agents to enhance the challenges for the following events. At the same time, AFT partners should keep working on trying to provide new ideas to foster cooperation and share knowledge and experiences on hot AFT topics like new crypto-threats, financing trends and terrorism *modus operandi*. In this way, it will be possible to create a more homogeneous community in which LEAs and FIUs can take inspiration for detecting and defining new investigation policies and share experiences about new market products useful for their investigations. The last future target would be to improve the number of LEAs and FIUs participants. In that sense, online participation can represent an interesting and valid solution by exploiting the pilot test led in The Hague events.

**References**

Anti-FinTer, 2023. *Versatile artificial intelligence investigative technologies for revealing online cross-border financing activities of terrorism.* s.l., s.n.

ASGARD, 2016. *Analysis System for Gathered Raw Data.* s.l., s.n.

Auman, C., 2011. Using simulation games to increase student and instructor engagement. *College Teaching,* Volume 59, p. 154–161.

Barrows, H., 2002. Is it truly possible to have such a thing as dPBL?. *Distance Education,* Volume 23, p. 119–122.

Bates, R., 2004. A critical analysis of evaluation practice: the Kirkpatrick model and the principle of beneficence. *Evaluation and program planning,* Volume 27, p. 341–347.

Carlile, O. & Jordan, A., 2005. It works in practice but will it work in theory? The theoretical underpinnings of pedagogy. *Emerging issues in the practice of university learning and teaching,* Volume 1, p. 11–26.

Chicone, R. G. a. F. S., 2020. A comparison study of two cybersecurity learning systems: facebook's open-source capture the flag and CTFd. *Issues in Information Systems,* Volume 21, pp. 202--212.

CTC, 2023. *Cut The Cord.* s.l., s.n.

CYCLOPES, 2023. *Cybercrime Law Enforcement Practitioners Networks.* s.l., s.n.

DANTE, 2018. *Detecting and Analysing Terrorist-Related Online Contents and Financing Activities.* s.l., s.n.

Eagle, C. & Clark, J. L., 2004. *Capture-the-flag: Learning computer security under fire,* s.l.: s.n.

ENISA, 2021. *All you need to know about Capture the Flag competitions.* s.l., s.n.

Europol, 2021. *Internet Organised Crime Threat Assessment (IOCTA).* s.l., s.n.

Europol, 2022. *European Union Terrorism Situation and Trend Report, Publications Office of the European Union.* s.l., s.n.

Finkel, D. L., 2000. Teaching with your mouth shut. *Education Review.*

Garrison, D. R. & Arbaugh, J. B., 2007. Researching the community of inquiry framework: Review, issues, and future directions. *The Internet and higher education,* Volume 10, p. 157–172.

Governors Western University, 2020. *What is humanistic learning theory in education?.* s.l., s.n.

GRACE, 2020. *Global Response Against Child Exploitation.* s.l., s.n.

Hanafi, A. H. A. a. R. H. a. I. A. D. a. I. Z.-A. a. Z. M. N. A. a. R. F. A., 2021. A CTF-Based Approach in Cyber Security Education for Secondary School Students. *electronic Journal of Computer Science and Information Technology,* Volume 7.

Haslhofer, B., Stütz, R., Romiti, M. & King, R., 2021. GraphSense: A General-Purpose Cryptoasset Analytics Platform. *Arxiv pre-print.*

Huang, H., Ding, J., Zhang, W. & Tomlin, C. J., 2011. *A differential game approach to planning in adversarial scenarios: A case study on capture-the-flag.* s.l., s.n., p. 1451–1456.

i-LEAD, 2023. *innovation - Law Enforcement Agency's Dialogue.* s.l., s.n.

Jiang, C., Foye, J., Broadhurst, R. & Ball, M., 2021. Illicit firearms and other weapons on darknet markets. *Trends and Issues in Crime and Criminal Justice [electronic resource],* p. 1–20.

Kilger, M. & Choo, K.-K. R., 2022. *Do Dark Web and Cryptocurrencies Empower Cybercriminals?.* s.l., s.n., p. 277.

Klingberg, S., 2022. Countering Terrorism: Digital Policing of Open Source Intelligence and Social Media Using Artificial Intelligence. In: *Artificial Intelligence and National Security.* s.l.:Springer, p. 101–111.

Maher, D., 2017. Can artificial intelligence help in the war on cybercrime?. *Computer Fraud & Security,* Volume 2017, p. 7–9.

McDaniel, L., Talvi, E. & Hay, B., 2016. *Capture the flag as cyber security introduction.* s.l., s.n., p. 5479–5486.

McLean, H., 2018. This is the way to teach: insights from academics and students about assessment that supports learning. *Assessment & Evaluation in Higher Education,* Volume 43, p. 1228–1240.

Mijwil, M. a. A. M. a. o., 2023. Towards artificial intelligence-based cybersecurity: the practices and ChatGPT generated ways to combat cybercrime. *Iraqi Journal For Computer Science and Mathematics,* Volume 4, pp. 65--70.

O'Neill, G. & McMahon, T., 2005. Student-centred learning: What does it mean for students and lecturers.

Peres, S. C., Pham, T. & Phillips, R., 2013. *Validation of the system usability scale (SUS) SUS in the wild.* s.l., s.n., p. 192–196.

Smidt, A., Balandin, S., Sigafoos, J. & Reed, V. A., 2009. The Kirkpatrick model: A useful tool for evaluating training outcomes. *Journal of Intellectual and Developmental Disability,* Volume 34, p. 266–274.

Švábenskỳ, V., Čeleda, P., Vykopal, J. & Brišáková, S., 2021. Cybersecurity knowledge and skills taught in capture the flag challenges. *Computers & Security,* Volume 102, p. 102154.

TITANIUM, 2019. *Tools for the Investigation of Transactions in Underground Markets.* s.l., s.n.

Topping, K., 1998. Peer assessment between students in colleges and universities. *Review of educational Research,* Volume 68, p. 249–276.

Zola, F., Eguimendia, M., Bruse, J. L. & Urrutia, R. O., 2019. *Cascading machine learning to attack bitcoin anonymity.* s.l., s.n., p. 10–17.